\documentclass[lettersize,journal]{IEEEtran}
\usepackage{amsmath,amsfonts}
\usepackage{amssymb}
\usepackage{graphicx}
\usepackage{algorithm}
\usepackage{array}
\usepackage[caption=false,font=normalsize,labelfont=sf,textfont=sf]{subfig}
\usepackage{textcomp}
\usepackage{stfloats}
\usepackage{url}
\usepackage{verbatim}
\usepackage{graphicx}
\usepackage{cite}
\usepackage{multirow}
\usepackage{multicol}
\usepackage{colortbl}
\usepackage[table]{xcolor}
\usepackage{threeparttable}
\usepackage{pifont}
\usepackage{circledsteps}
\usepackage{caption}
\usepackage{algpseudocode}

\algrenewcommand\textproc{}
\begin{document}

\title{An Experimental Study of SOTA LiDAR Segmentation Models}

\author{Bike Chen, Antti Tikanmäki, Juha Röning
}




\maketitle

\begin{abstract}
Point cloud segmentation (PCS) is to classify each point in point clouds. The task enables robots to parse their 3D surroundings and run autonomously. According to different point cloud representations, existing PCS models can be roughly divided into point-, voxel-, and range image-based models. However, no work has been found to report comprehensive comparisons among the state-of-the-art point-, voxel-, and range image-based models from an application perspective, bringing difficulty in utilizing these models for real-world scenarios. In this paper, we provide thorough comparisons among the models by considering the LiDAR data motion compensation and the metrics of model parameters, max GPU memory allocated during testing, inference latency, frames per second, intersection-over-union (IoU) and mean IoU (mIoU) scores. The experimental results benefit engineers when choosing a reasonable PCS model for an application and inspire researchers in the PCS field to design more practical models for a real-world scenario.
\end{abstract}


\section{Introduction}\label{sec:intro}
\IEEEPARstart{P}{oint} cloud segmentation (PCS) is to classify each point into one of the predefined classes. This task plays a significant role in robotics. For example, PCS enables robots to search free spaces and avoid obstacles~\cite{semantickitti_2019_behley,semanticposs_2020,nuscenes_panoptic,domainAdpt2024}. Besides, the semantic predictions of PCS models can be utilized to build a semantic map~\cite{sa_loam_2021, suma++_2019} for robotic navigation. 

\begin{figure*}[t]
	\centering
	\includegraphics[width=1.6\columnwidth]{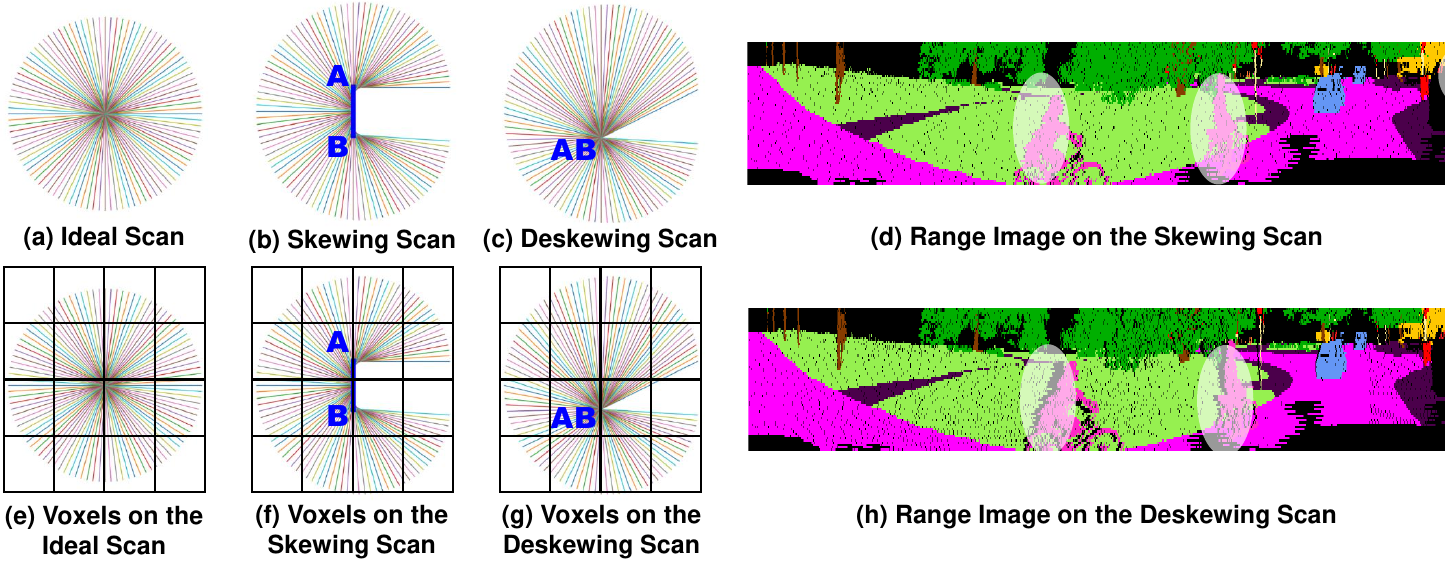}
	\caption{(a) is an ideal scan (\textit{i.e.}, only show the top view). When the LiDAR sensor does not move, the collected raw points form a circle. (b) is a skewing scan. When the LiDAR sensor moves from A to B quickly, the points are distorted. (c) is a deskewing scan. All points are aligned at the end of the scan with motion compensation. (e), (f) and (g) are voxels on the above (a), (b), and (c), respectively (\textit{i.e.}, only show the top view). (d) is the range image on the skewing scan. (h) is the range image on the deskewing scan. There are some black holes around the objects in (h) (\textit{i.e.}, see the white masks).}
	\label{fig:motion_compensation}  
\end{figure*}

In this paper, we study the point-, voxel-, and range image-based models on outdoor point cloud datasets such as SemanticKITTI~\cite{semantickitti_2019_behley} and nuScenes~\cite{nuscenes_panoptic}. A point cloud can be represented by points, voxels, and a range image. Correspondingly, the PCS models are roughly grouped into point-based methods~\cite{pointnet++_2017,kpconv_2019}, voxel-based approaches~\cite{minkowski2019,spvnas_2020,spherical_transformer_2023}, and range image-based solutions~\cite{fidnet_2021,cenet_2022,filling_missing2024,pdm2024}. The development of advanced PCS models has attracted much attention.

However, to the best of our knowledge, no paper has been found to provide thorough comparisons among these types of models on the point cloud datasets from an application perspective, bringing difficulty in choosing the models for applications such as robotics and autonomous vehicles. In the applications, the following key factors should be considered. (1) Motion compensation~\cite{loam2014,kissicp2023} corrects the distorted points in the point cloud. Specifically, the LiDAR sensor is mounted on a robot and set to a fixed revolving speed, such as 10 Hz or 20 Hz. However, when the robot moves, the captured raw points will not form an ideal circle from the top view. We call the captured point cloud a skewing or distorted scan (see Fig.~\ref{fig:motion_compensation}(b)). Specifically, in Fig.~\ref{fig:motion_compensation}(b), the measured ranges of the objects with respect to A have been changed when the LiDAR moves to B in the whole scan. Hence, motion compensation is required to correct the distorted points by commonly aligning all points at the end of the scan (see Fig.~\ref{fig:motion_compensation}(c)). The corrected point cloud is called a deskewing or undistorted scan.

The important thing is that the skewing or deskewing scan has a different impact on the voxel and range image representations. Figs.~\ref{fig:motion_compensation}(f) and (g) show the voxels on the skewing and deskewing scans from the top view, respectively. Figs.~\ref{fig:motion_compensation}(d) and (h) depict the range images on the skewing and deskewing scans, respectively. We see that there are black holes around the objects in the range image produced on the deskewing scans.

(2) Other factors, such as the number of model parameters, max GPU memory allocated during inference, inference latency, frames per second (FPS), intersection-over-union (IoU) scores over each class, and mean IoU (mIoU) scores over all classes, should be considered together. For example, the latency or FPS of the models during testing must be taken into consideration in the real-time applications, although obtaining high IoU and mIoU scores in computer vision fields is a de facto decisive factor in paper acceptance~\cite{openpcseg2023,taseg24,rapid25}. Therefore, a comprehensive comparison study of the PCS models from an application perspective is required. 

In addition, a recent work~\cite{empir24} is similar to this paper. It provides comprehensive comparisons among various state-of-the-art (SOTA) PCS models with advanced data augmentation techniques~\cite{polarmix_2022,lasermix_2023} and sparse convolution engines~\cite{minkowski2019,spconv2022,torchsparse23}. However, the work~\cite{empir24} does not consider the key factor of motion compensation. Also, the comparison results among different models are fragmented, challenging users to directly analyze the results under the number of model parameters, max GPU memory allocated, inference latency, FPS, and the mIoU scores.

To provide thorough and fair comparisons among existing SOTA PCS models from an application view, we conduct experiments with the following steps: (1) All PCS models are trained with the same data augmentation combination on the skewing and deskewing scans, respectively. (2) The point-, voxel-, and range image-based models with high mIoU scores are chosen for further comprehensive comparisons. Extensive experiments are conducted on the popular SemanticKITTI~\cite{semantickitti_2019_behley} and nuScenes~\cite{nuscenes_panoptic} datasets. We believe that the experimental results and analysis benefit engineers and researchers in the community.

\section{Related Work}\label{sec:related_works}
In the section, we first review the associated point-, voxel- and range image-based approaches and data augmentation techniques. Then, we review the works that are similar to ours. Here, we roughly group existing point cloud segmentation (PCS) methods into the above three categories because of the point cloud representations and the proportion of the point-, voxel-, or range image-based operations in the architectures. 

\subsection{Point-Based Approaches}
Point-based models are built with various point-based operations. The key in the models is to build effective feature extraction modules to extract semantic features over a set of points. The pioneering work, PointNet~\cite{pointnet++_2017}, uses MLP (multi-layer perceptrons) layers to extra per-point features and adopts global aggregation operations, such as pooling, to effectively build semantic relationships among points to make the final per-point predictions. However, PointNet fails to build an effective local feature extraction component in the architecture. To overcome the drawback, the following works, such as PointNet++~\cite{pointnet++_2017}, PCNN~\cite{pcnn2018}, DGCNN~\cite{dgcnn2019}, KPConv~\cite{kpconv_2019}, and Point Transformer~\cite{pointtrans2021}, and KPConvX~\cite{kpconvx24}, focus on the designs of the local feature extraction modules. However, due to the sparsity and disorder properties of the points, incorporating too many complex point-based operations into the architecture inevitably leads to heavy computational overhead, hindering the application of the models on large-scale outdoor point clouds. To reduce the computational cost, the recent work, WaffleIron~\cite{waffleiron23}, introduces a waffle iron (WI) block in the architecture, where all points are projected onto the XY, XZ, and YZ planes sequentially, and 2D convolution operations are employed on the planes to extract features. Besides, the work, Point Transformer V3~\cite{pointtransv32024}, first makes the unordered points ordered by serializing all points with space-filling curves, such as the Z-order and Hilbert curves. Then, a transformer architecture is applied to the ordered points. In this paper, we choose WaffleIron to conduct comparison experiments because it is open-source and can run faster than other point-based models on the large-scale outdoor point clouds.

\subsection{Voxel-Based Methods}
Here, we only briefly review the sparse convolution-based models because the models can avoid the high computational cost and huge memory load by doing convolution operations only on the occupied voxels. The typical work is MinkowskiNet~\cite{minkowski2019} (or MinkUNet), built by generalized sparse convolutions. Many of the following works adopt MinkUNet as the backbone. For example, SPVCNN~\cite{spvnas_2020} uses MinkUNet as the main branch with a complementary point-based branch. UniSeg~\cite{openpcseg2023} adopts MinkUNet as the point-voxel backbone. TASeg~\cite{taseg24} and RAPid-Seg~\cite{rapid25} are built on the basic MinkUNet architecture. The above models adopt the classic cubic partition of the large-scale outdoor point clouds as the input but do not consider the point distribution. To cope with the problem, the works Cylinder3D~\cite{cylindrical3d2021} and SphereFormer~\cite{spherical_transformer_2023} introduce cylindrical partition and radial window partition, respectively. In this paper, we choose MinkUNet and SPVCNN to conduct comparison experiments because, among the above voxel-based models, MinkUNet and SPVCNN can achieve state-of-the-art performance regarding the speed-accuracy trade-off.

\subsection{Range Image-Based Solutions}
Typically, the backbones of the range image-based PCS models are usually derived from 2D image classification or segmentation backbones. Besides, to avoid the loss of information caused by the projection process, a point-based post-processing component is commonly used to refine the final pointwise predictions. For example, the pioneering work, SqueezeSeg~\cite{squeezeseg}, uses modified SqueezeNet~\cite{squeezenet_2016} to make per-point predictions. Also, a recurrent CRF (conditional random field) component is adopted to refine the predictions further. Another classic work, RangeNet++~\cite{rangenet++}, adopts revised DarkNet~\cite{yolov3_2018} first to make per-pixel predictions. Then, a range image-based $K$NN ($K$-nearest neighbor search) module is introduced to refine the image-to-points predictions efficiently. The following work, CENet~\cite{cenet_2022}, utilizes a modified ResNet~\cite{resnet_2016} as the backbone and uses the $K$NN to make the final pointwise predictions. In FIDNet~\cite{fidnet_2021}, the authors also use the revised ResNet as the backbone, but they propose a range image-based NLA (nearest label assignment) module to refine the final per-point predictions. RangeFormer~\cite{rangeformer_2023} utilizes the SegFormer-like~\cite{segformer2021} backbone to make the pointwise predictions on each subset of the point cloud, and then the predictions from several sub-clouds are concatenated and further refined by the proposed RangePost and $K$NN. RangeViT~\cite{rangevit_2023} opts the revised ViT~\cite{vit_iclr_2021} as the backbone and adopts KPConv~\cite{kpconv_2019} to make the final per-point predictions. Recent Fast FMVNet~\cite{filling_missing2024} and Fast FMVNet V2~\cite{pdm2024} adopt the modified ConvNeXt-Tiny~\cite{convnext2022} as the backbone and utilize NLA and the proposed PDM (pointwise decoder module) to refine the final predictions, respectively. In this paper, we choose CENet and Fast FMVNet V2 to conduct comparison experiments because both models can run at high speed and achieve state-of-the-art performance.

\subsection{Data Augmentation Techniques}
To effectively train the PCS models, diverse point cloud samples are required and can be created by the combination of various data augmentation techniques. Here, we split the data augmentation techniques into two groups: basic and spinning LiDAR-specific data augmentation techniques. The basic augmentation techniques~\cite{data_aug2024} include random flipping, random scaling, random rotation, random dropping, and random jittering. These techniques can be applied to any point cloud data, such as solid-state LiDAR-based data and mechanical spinning LiDAR-based data. By contrast, spinning LiDAR-specific augmentation techniques can commonly be employed on the spinning LiDAR-based datasets, such as SemanticKITTI~\cite{semantickitti_2019_behley} and nuScenes~\cite{nuscenes_panoptic}. The specific augmentation techniques includes PolarMix~\cite{polarmix_2022}, LaserMix~\cite{lasermix_2023}, and VRCrop~\cite{pdm2024}. PolarMix swaps the sub-clouds between two point clouds along the horizontal direction, while LaserMix swaps the sub-clouds along the vertical direction. Also, PolarMix and LaserMix include rare object augmentation to make class balance during training. In practice, both augmentation techniques are usually used together~\cite{rangeformer_2023}. VRCrop, instead of the traditional copy-rotate-paste augmentation for the point clouds, introduces the concept of a virtual range image where the point cloud is reorganized and looks like a 2D image. Hence, the point cloud-based copy-rotate-paste augmentation can be like the color image-based copy-paste augmentation. One advantage of VRCrop is the elimination of undesirable artifacts in the augmented point cloud. In this paper, we provide a combination of data augmentation techniques. Under the combination, existing PCS models are trained again for fair comparisons.

\subsection{Works Focusing on Comparison Experiments}
All PCS works focusing on performance improvement provide comprehensive comparisons between existing approaches and the proposed method. For example, the work, UniSeg~\cite{openpcseg2023}, compares the proposed models (UniSeg $0.2\times$ and UniSeg $1.0\times$) with many existing PCS models, such as MinkowskiNet and SPVCNN, to show the superior performance in terms of the number of model parameters, inference latency, and mIoU scores. However, the mIoU scores of the compared models are outdated, leading to unfair comparisons. In addition, the work~\cite{empir24}, focusing on how to train the existing state-of-the-art PCS models, provides the comparison results among various PCS models with the above metrics. However, the comprehensive comparison results are only shown in the figure. More importantly, these works ignore the impact of motion compensation on the model performance. Different from the previous works, this paper first considers the effect of motion compensation and trains most PCS models again to reach state-of-the-art performance. Then, comprehensive comparisons among the models are reported with the metrics of the number of model parameters, max GPU memory allocated during inference, inference latency, frames per second, IoU and mIoU scores.

\section{Models}
This section describes the motion compensation, point cloud segmentation (PCS) models, and the data augmentation combination.

\subsection{Motion Compensation}
Motion compensation aims to correct skewing point clouds that are directly output from the LiDAR sensor. Skewing and deskewing point clouds have different impacts on different point cloud segmentation models. Besides, in the application, a PCS model developed on the skewing point clouds can be directly employed. By contrast, the PCS model based on the deskewing point clouds must be applied after the raw point clouds are corrected with motion compensation. Therefore, different PCS models correspond to different application stages.

In motion compensation, we assume that for a point cloud, the angular velocity $\boldsymbol{w}_t \in \mathbb{R}^3$ and translation velocity $\boldsymbol{v}_t \in \mathbb{R}^3$ at the time $t$ have been obtained by a SLAM system~\cite{loam2014,kissicp2023} or an inertial measurement unit (IMU)~\cite{fastlio2}. Besides, for a skewing point $\boldsymbol{p} \in \mathbb{R}^3$, the sampling timestamp relative to that of the first point is $\alpha \in \left[0, \triangle t\right]$ where $\triangle t$ indicates the acquisition time of the whole point cloud. Therefore, the deskewing point $\boldsymbol{p}^{\ast}$ can be calculated by the following Eq.~(\ref{eq:deskewing_point}), which is:
\begin{equation}\small\label{eq:deskewing_point}
	\boldsymbol{p}^{\ast} = \text{Exp}(\alpha\boldsymbol{w}_t)\boldsymbol{p} + \alpha\boldsymbol{v}_t,
\end{equation}
where $\text{Exp}: \mathbb{R}^3 \rightarrow SO(3)$ means the transformation from the tangent space to the manifold space. Note that we do not include the time of motion compensation in the comprehensive comparisons to make the results consistent with those in the existing works. 

\begin{figure}
	\centering
	\includegraphics[width=0.95\columnwidth]{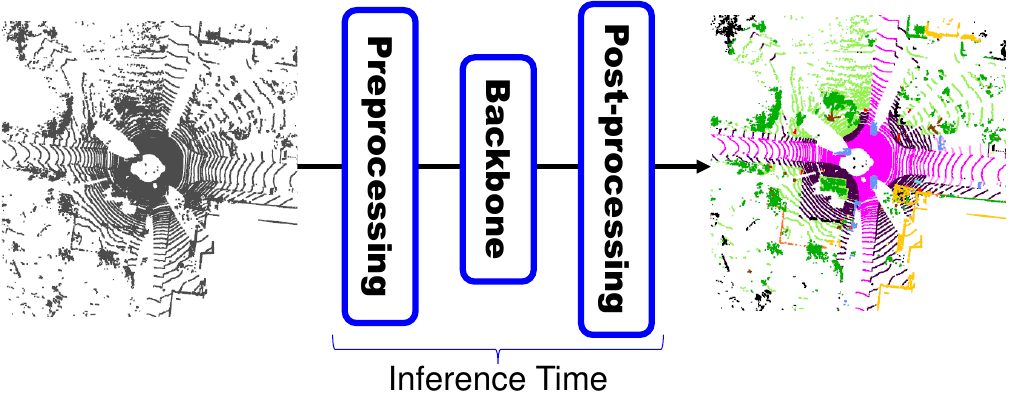}
	\caption{The pipeline of the point cloud segmentation (PCS) models. }
	\label{fig:pipeline}  
\end{figure}

\subsection{Point Cloud Segmentation Models}
In this paper, we choose WaffleIron~\cite{waffleiron23}, MinkUNet~\cite{minkowski2019}, SPVCNN~\cite{spvnas_2020}, CENet~\cite{cenet_2022}, and Fast FMVNet V2~\cite{pdm2024} to conduct comparison experiments, because these models can achieve the state-of-the-art performance in terms of the speed-accuracy trade-off. 

These models can be briefly described by the pipeline in Fig.~\ref{fig:pipeline}. Specifically, (1) in the preprocessing step, the point-based model WaffleIron downsamples the point cloud. The voxel-based models MinkUNet and SPVCNN convert the points into voxels. The range image-based models CENet~\footnote{We use the CENet+PDM in the work~\cite{pdm2024}} and Fast FMVNet V2 project the points onto the range images~\cite{filling_missing2024}\footnote{Scan unfolding++ is adopted to prepare range images.}. (2) The backbone means the architecture of each model. (3) In the post-processing step, WaffleIron adopts nearest neighbor interpolation to propagate the predictions to all points. MinkUnet and SPVCNN directly map the voxel-based predictions to all points. CENet and Fast FMVNet V2 utilize the pointwise decoder module~\cite{pdm2024} to project the predictions back to all points and further refine the final per-point predictions. 

In addition, for fair comparisons, the times of preprocessing and post-processing are included in the inference latency, although part of the preprocessing and post-processing time can overlap with the backbone inference time by a programming trick.

\subsection{Data Augmentation Combination}
Data augmentation combination aims to create diverse point cloud samples on which models are trained to reach state-of-the-art performance in terms of the mIoU scores. To ensure fair comparisons, we use the same data augmentation combination for the above point cloud segmentation (PCS) models during training. 

\begin{figure}[t]
	\centering
	\includegraphics[width=0.55\columnwidth]{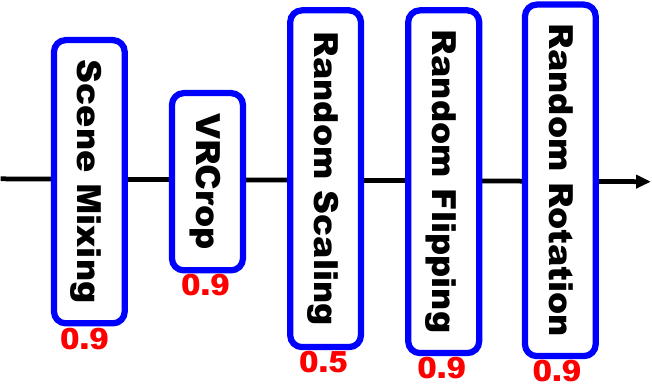}
	\caption{The data augmentation combination used in the paper. The red fonts indicate the probabilities during training.}
	\label{fig:data_augmentation_combo}  
\end{figure}

The data augmentation combination is depicted in Fig.~\ref{fig:data_augmentation_combo}, and it contains \textit{Scene Mixing}~\cite{polarmix_2022,lasermix_2023}, \textit{VRCrop}~\cite{pdm2024}, \textit{Random Scaling}, \textit{Random Flipping}, and \textit{Random Rotation}. Specifically, \textit{Scene Mixing} is the combination of the scene mixing parts in PolarMix~\cite{polarmix_2022} and LaserMix~\cite{lasermix_2023}. The settings are provided in Table~\ref{tab:scene_mixing}. \textit{VRCrop}~\cite{pdm2024} is the upgraded copy-rotate-paste augmentation targeting the point clouds. \textit{VRCrop} can avoid the undesirable artifacts in the augmented point clouds for the range image-based models. \textit{Random Scaling}, \textit{Random Flipping}, and \textit{Random Rotation} means that the points are scaled, flipped, and rotated, respectively. 

\begin{table}[t]
	\caption{Scene Mixing contains PolarMix~\cite{polarmix_2022} and LaserMix~\cite{lasermix_2023} in the data augmentation combination. In (A, B), A and B mean the number of row and column patches.}
	\label{tab:scene_mixing}
	\centering
	\scalebox{1.0}{
		\begin{tabular}{l| c|c}
			&	SemanticKITTI                             & nuScenes        \\ \hline 
			LaserMix~\cite{lasermix_2023}   &	(2, 1), (3, 1), (4, 1), (5, 1), (6, 1)    & (2, 1), (3, 1)  \\ 
			PolarMix~\cite{polarmix_2022}   &	(1, 2), (1, 2), (1, 2), (1, 2), (1, 2)    & (1, 2), (1, 2)  \\
	\end{tabular}}
\end{table}

Note that in experiments, without VRCrop means the traditional copy-rotate-paste augmentation adopted (\textit{i.e.}, instance-level rotate-pasting in PolarMix~\cite{polarmix_2022}). Besides, we explore \textit{VRCrop} in training the point- and voxel-based models in this paper because \textit{VRCrop} was only utilized for range image-based models~\cite{pdm2024} before.

\section{Experiments}
In the section, we first briefly describe the experimental settings. Then, we show the impacts of motion compensation and VRCrop~\cite{pdm2024} on model performance. Finally, we report the comprehensive comparison results on the SemanticKITTI~\cite{semantickitti_2019_behley} and nuScenes~\cite{nuscenes_panoptic} datasets.
 
\subsection{Experimental Settings}
In the subsection, we describe the datasets and training and testing settings.

\subsubsection{Datasets}
SemanticKITTI~\cite{semantickitti_2019_behley} and nuScenes~\cite{nuscenes_panoptic} are used to conduct all experiments. Both datasets are collected by the mechanical spinning LiDAR sensors. Specifically, SemanticKITTI~\cite{semantickitti_2019_behley} data was captured by a 64-beam LiDAR sensor with 10 Hz, while nuScenes~\cite{nuscenes_panoptic} data was collected by a 32-beam LiDAR sensor with 20 Hz. Hence, the point clouds in SemanticKITTI are sparser than that in nuScenes. 

In SemanticKITTI, the sequences ``$00\sim07$ and $09\sim10$", ``$08$", and ``$11\sim21$" are used as the training, validation, and testing datasets, respectively. Besides, 19 classes are taken into consideration under the single scan condition. Note that we use the approach in the work~\cite{filling_missing2024} to recover all ``skewing" point clouds to carry out experiments because of no raw point clouds for the sequence $03$ and all testing sequences. In nuScenes, there are 28,130 training, 6,019 validation, and 6,008 testing samples, respectively. Also, only 16 classes are considered in the experiments.

\subsubsection{Training \& Testing Settings}

\begin{table}[t]
	\caption{The experimental settings during training. ``P": point-based model, namely WaffleIron~\cite{waffleiron23}; ``V": voxel-based models, namely MinkUNet~\cite{minkowski2019} and SPVCNN~\cite{spvnas_2020}; ``I": range image-based models, namely CENet~\cite{cenet_2022} and Fast FMVNet V2~\cite{pdm2024}.}
	\label{tab:experimental_settings}
	\centering
	\scalebox{1.0}{
		\begin{tabular}{l|c|c}
			& SemanticKITTI      & nuScenes         \\ \hline 
			Optimizer	    & \multicolumn{2}{c}{\textit{AdamW}}    \\  
			Base Learning Rate   & \multicolumn{2}{c}{0.002}	    \\
			Weight Decay    & \multicolumn{2}{c}{0.0001}            \\
			Training Epochs & 45 (P), 50 (V/I)   & 45 (P), 80 (V/I) \\
			Warm-up Epochs  & 4 (P), 10 (V/I)    & 9 (P), 20 (V/I)  \\
			Warm-up Type    & \multicolumn{2}{c}{\textit{cos}}      \\
			Decay Type      & \multicolumn{2}{c}{\textit{exp}}      \\
			Batch Sizes     &  8 (P/I), 16 (V)   & 8 (P/I), 16 (V)  \\
			All Random Seeds    & \multicolumn{2}{c}{\textit{123}}      \\
	\end{tabular}}
\end{table}

All models are trained on four Nvidia A100 GPUs and tested on one Nvidia A100 GPU. The detailed experimental settings are provided in Table~\ref{tab:experimental_settings}. Note that all random seeds are fixed for reproduction during training and testing. Moreover, no test-time augmentation and ensemble techniques are utilized to boost the final performance further because these techniques cause the significantly increased inference time~\cite{empir24}.

\subsection{Experimental Results}
In this section, we first study the effects of motion compensation and VRCrop~\cite{pdm2024} on the point- and voxel-based models, because both techniques are only studied on the range image-based models. Then, we provide thorough comparison results on the SemanticKITTI and nuScenes datasets.

\begin{table}[t]
	\caption{Comparison results among the models trained on skewing and deskewing point clouds of the SemanticKITTI~\cite{semantickitti_2019_behley} dataset with and without VRCrop. ``\ding{52}": with skewing point clouds or with VRCrop.}
	\label{tab:deskewing_vrcrop}
	\centering
	\scalebox{1.0}{
		\begin{tabular}{l|c|c|c}
			Models                                       & Skewing   & VRCrop    & mIoU (\%)$\uparrow$   \\ \hline 
			\multirow{5}*{WaffleIron~\cite{waffleiron23}}&           &           & \cellcolor{gray!40}{68.03}  \\
			                                             &           &           & \textbf{69.00} \\
														 & \ding{52} &           & 68.40 \\
			                                             &           & \ding{52} & 68.55 \\ 
			                                             & \ding{52} & \ding{52} & 68.53 \\ \hline
			                                             
            \multirow{5}*{MinkUNet (cr1.6)~\cite{minkowski2019}} &   &           & \cellcolor{gray!40}{70.04}  \\
													     &           &           & 68.71 \\
													     & \ding{52} &           & 68.52 \\ 
			                                             &           & \ding{52} & \textbf{69.41} \\ 
			                                             & \ding{52} & \ding{52} & 68.74 \\ \hline
			                                             
		    \multirow{5}*{SPVCNN (cr1.6)~\cite{spvnas_2020}} &       &           & \cellcolor{gray!40}{68.58}  \\
		                                                 &           &           & 68.58  \\
		                                                 & \ding{52} &           & 68.44  \\
		                                                 &           & \ding{52} & \textbf{69.07}  \\
		                                                 & \ding{52} & \ding{52} & 68.46  \\ \hline                                        
			
			\multirow{2}*{CENet~\cite{pdm2024}}          & \ding{52} &           & \cellcolor{gray!40}{64.64} \\ 
			                                             & \ding{52} & \ding{52} & \cellcolor{gray!40}{\textbf{66.33}} \\ \hline
			
			\multirow{2}*{Fast FMVNet V2~\cite{pdm2024}} & \ding{52} &           & \cellcolor{gray!40}{66.03} \\ 
			                                             & \ding{52} & \ding{52} & \cellcolor{gray!40}{\textbf{68.55}} \\ 
	\end{tabular}}
\end{table}

\subsubsection{Impacts of Motion Compensation and VRCrop}
For fair comparisons, we reproduce the experimental results with the above experimental settings. For WaffleIron~\cite{waffleiron23}, the mIoU score 68.03\% is obtained by testing the pre-trained model and open-source code~\footnote{WaffleIron: https://github.com/valeoai/WaffleIron}. The score 69.0\% is the reproduced result. For MinkUNet~\cite{minkowski2019} and SPVCNN~\cite{spvnas_2020}, the mIoU scores, \textit{i.e.}, 70.04\% and 68.58\%, are copied from the OpenPCSeg~\footnote{OpenPCSeg: https://github.com/PJLab-ADG/OpenPCSeg}. The reproduced results are 68.71\% and 68.58\%, respectively. Note that according to the work~\cite{empir24}, the final mIoU scores of MinkUNet and SPVCNN with the TorchSparse~\cite{torchsparse23} backend might fluctuate by about 1.5\% for various random seeds. Therefore, the reproduced results are still reasonable. Besides, we provide the mIoU scores of the range image-based models CENet and Fast FMVNet V2 for comparison analysis. The results are copied from the work~\cite{pdm2024}.

In Table~\ref{tab:deskewing_vrcrop}, we see that the skewing point clouds and VRCrop have negative impacts on the performance of the point-based model WaffleIron. The potential explanations are as follows: (1) WI blocks in WaffleIron project points onto XY, XZ, and YZ planes sequentially. The blocks might require undistorted objects during projection. However, skewing LiDAR data leads to distorted objects, thereby leading to degenerated performance. (2) VRCrop avoids object occlusion along the LiDAR beam direction by deleting the conflicting points in the augmented point cloud. However, object occlusion might not be a serious problem in WaffleIron because WI blocks project points onto the XY plane (Bird's Eye View (BEV)), and all objects in this plane are separated. Also, the conclusion in the work~\cite{waffleiron23}, namely ``\textit{projecting only in the XY plane (BEV) has a negligible impact on the mIoU scores}", can support the explanation. (3) Besides, during training, a fixed number of points within a local area are chosen as the input, so the input (\textit{i.e.}, without VRCrop) with more points of the rare objects benefits the performance improvement.

For the voxel-based models, namely MinkUNet and SPVCNN, the skewing point clouds lead to inferior models' performance, while VRCrop increases the mIoU scores of both models. The potential explanations are as follows: (1) Similar to the aforementioned explanation, voxel-based models require undistorted objects in the 3D space. (2) Different from the training setting in WaffleIron, the whole point cloud acts as the input in training MinkUNet and SPVCNN. In this situation, VRCrop enables better class balance than the traditional copy-rotate-paste augmentation technique for creating the diverse point cloud.

For the range image-based models, namely CENet and Fast FMVNet V2, the works~\cite{filling_missing2024,pdm2024} have shown the positive impacts of the skewing point clouds and VRCrop on the performance. Here, we provide the explanations again: (1) The range image on the deskewing point cloud has black holes around the objects (see Fig.~\ref{fig:motion_compensation}(h)), damaging the shapes and bringing difficulty in training the models. (2) VRCrop avoids object occlusion and undesirable artifacts in the augmented point cloud, so VRCrop can eliminate the semantic misalignment between the ``range image-to-points" outputs and the ground truth during training.

\subsubsection{Comprehensive Comparisons on SemanticKITTI}
\begin{figure}[t]
	\centering
	\includegraphics[width=0.98\columnwidth]{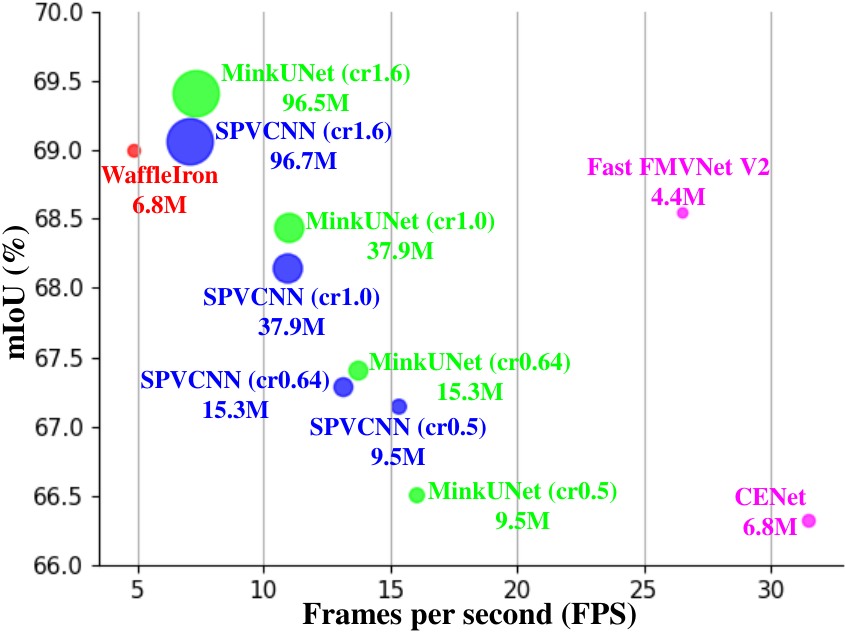}
	\caption{Comparison results among point-, voxel-, and range image-based models~\cite{waffleiron23,minkowski2019,spvnas_2020,pdm2024} on the SemanticKITTI~\cite{semantickitti_2019_behley} validation dataset in terms of the model parameters (M), frames per second (FPS), mIoU scores (\%). The sizes of the circles indicate the number of model parameters.}
	\label{fig:speed_miou_kitti}
\end{figure}

The comprehensive comparison results on the SemanticKITTI~\cite{semantickitti_2019_behley} are provided in Tables~\ref{tab:time_kitti_models} and~\ref{tab:64x2048_kitti_val_results} and Fig.~\ref{fig:speed_miou_kitti}. 

\begin{table*}[t]
	\caption{Comparison results on the SemanticKITTI~\cite{semantickitti_2019_behley} validation dataset in terms of IoU and mIoU scores (\%). \textbf{Bold font}: the best result; \underline{Underline}: the second-best result.}
	\label{tab:64x2048_kitti_val_results}
	\centering
	\scalebox{0.85}{
		\begin{tabular}{l|c|c|c|c|c|c|c|c|c|c|c|c|c|c|c|c|c|c|c|c}
			Models	 & mIoU &\rotatebox{90}{Car} &\rotatebox{90}{Bicycle} &\rotatebox{90}{Motorcycle} &\rotatebox{90}{Truck} &\rotatebox{90}{Other-vehicle} &\rotatebox{90}{Person} &\rotatebox{90}{Bicyclist} &\rotatebox{90}{Motorcyclist} &\rotatebox{90}{Road} &\rotatebox{90}{Parking} &\rotatebox{90}{Sidewalk} &\rotatebox{90}{Other-ground} &\rotatebox{90}{Building} &\rotatebox{90}{Fence} &\rotatebox{90}{Vegetation} &\rotatebox{90}{Trunk} &\rotatebox{90}{Terrain} &\rotatebox{90}{Pole} &\rotatebox{90}{Traffic-sign}  \\ \hline 
			
			WaffleIron (c256) &69.0 &95.9 &57.7 &81.4 &89.1 &54.4 &\textbf{82.3} &91.4 &\textbf{0.8} &94.9 &50.1 &\underline{83.0} &\textbf{16.9} &90.8 &\underline{64.7} &\textbf{90.0} &\textbf{73.9} &\textbf{78.9} &\textbf{62.7} &\underline{52.3} \\ \hline
			
			MinkUNet (cr1.6)  &\textbf{69.4} &\textbf{97.6} &\textbf{59.8} &\textbf{84.0} &\underline{90.6} &\textbf{82.5} &79.3 &\underline{91.9} &0.0 &94.1 &52.4 &81.9 &14.2 &91.1 &62.3 &87.9 &68.1 &74.5 &58.4 &47.9 \\
			MinkUNet (cr1.0)  &68.4 &\underline{97.5} &\underline{58.7} &\underline{83.4} &89.0 &81.2 &77.9 &91.4 &0.0 &94.2 &51.7 &82.0 &7.7  &\underline{91.2} &62.0 &87.6 &66.9 &74.1 &57.8 &46.2 \\
			MinkUNet (cr0.64) &67.4 &97.5 &56.6 &77.4 &89.6 &79.7 &76.9 &91.5 &0.0 &94.0 &50.5 &81.3 &7.1  &90.5 &59.2 &88.1 &66.2 &75.3 &55.5 &44.0 \\
			MinkUNet (cr0.5)  &66.5 &96.9 &51.1 &79.7 &88.4 &71.6 &73.3 &90.8 &\underline{0.3} &93.6 &52.1 &81.1 &8.4  &90.4 &57.6 &88.3 &66.0 &76.3 &54.9 &43.1 \\ \hline
			
			SPVCNN (cr1.6)    &\underline{69.1} &97.5 &57.6 &80.9 &\textbf{92.7} &\underline{81.3} &80.0 &\textbf{92.5} &0.0 &93.8 &\underline{54.7} &81.7 &8.2  &\textbf{91.4} &62.9 &88.2 &68.6 &75.4 &58.4 &46.4 \\
			SPVCNN (cr1.0)    &68.1 &97.3 &58.1 &81.2 &89.3 &77.6 &77.8 &91.7 &0.2 &93.8 &51.0 &81.3 &9.6  &90.7 &59.8 &88.0 &68.0 &74.7 &57.4 &47.3 \\
			SPVCNN (cr0.64)   &67.3 &97.2 &56.5 &78.3 &86.0 &77.4 &77.1 &91.5 &0.2 &94.3 &51.2 &81.6 &7.6  &90.8 &59.4 &87.2 &67.2 &73.5 &55.9 &45.7 \\
			SPVCNN (cr0.5)    &67.2 &97.2 &54.9 &78.2 &90.1 &74.5 &75.7 &90.6 &0.0 &93.7 &51.8 &81.4 &7.5  &90.7 &59.7 &\underline{88.4} &66.4 &\underline{76.4} &54.3 &44.3 \\ \hline
			
			CENet             &66.3 &94.1 &50.5 &72.0 &83.2 &56.8 &\underline{81.8} &91.5 &0.0 &\underline{94.9} &41.8 &81.8 &13.5 &90.7 &59.7 &87.7 &\underline{70.4} &75.4 &60.4 &\textbf{54.0} \\
			Fast FMVNet V2    &68.5 &96.6 &46.5 &76.0 &87.6 &66.5 &81.4 &91.0 &0.0 &\textbf{95.8} &\textbf{57.4} &\textbf{84.1} &\underline{16.6} &90.8 &\textbf{65.6} &87.9 &69.3 &75.6 &\underline{61.5} &52.0  \\ 
	\end{tabular}}
\end{table*}

\begin{table}[t]
	\caption{Comparison results among the point cloud segmentation models~\cite{waffleiron23,minkowski2019,spvnas_2020,pdm2024} on the SemanticKITTI~\cite{semantickitti_2019_behley} validation dataset in terms of model parameters (Params.), max GPU memory allocated (Mem.), inference latency, frames per second (FPS), and mIoU scores (\%).}
	\label{tab:time_kitti_models}
	\centering
	\scalebox{0.945}{
		\begin{tabular}{l|c|c|c|c|c}
			Methods           &Params. & Mem.    & Latency &FPS   & mIoU (\%)  \\ \hline 
			WaffleIron (c256) & \cellcolor{blue!40}{6.8M}   & 3.22GB  & 207.3ms & \cellcolor{blue!40}{4.8}  & \cellcolor{blue!40}{69.00} \\ \hline
			
			MinkUNet (cr1.6)  & 96.5M  & 1.30GB  & 136.3ms & 7.3  & \textbf{69.41} \\   
			MinkUNet (cr1.0)  & 37.9M  & 0.76GB  & 90.6ms  & 11.0 & 68.44 \\  
			MinkUNet (cr0.64) & 15.3M  & 0.48GB  & 73.2ms  & 13.7 & 67.41 \\
			MinkUNet (cr0.5)  & \cellcolor{blue!40}{9.5M}   & \textbf{0.39GB}  & 62.6ms  & \cellcolor{blue!40}{16.0} & \cellcolor{blue!40}{66.51} \\ \hline
			
			SPVCNN (cr1.6)    & 96.7M  & 1.30GB  & 141.8ms & 7.1  & 69.06 \\ 
			SPVCNN (cr1.0)    & 37.9M  & 0.76GB  & 91.3ms  & 10.9 & 68.14 \\ 
			SPVCNN (cr0.64)   & 15.3M  & 0.49GB  & 76.5ms  & 13.1 & 67.29 \\ 
			SPVCNN (cr0.5)    & \cellcolor{blue!40}{9.5M}   & 0.39GB  & 65.4ms  & \cellcolor{blue!40}{15.3} & \cellcolor{blue!40}{67.15} \\ \hline
			
			CENet             & \cellcolor{blue!40}{6.8M}   & 2.27GB  & \textbf{31.8ms}  &\cellcolor{blue!40}{\textbf{31.5}}  & \cellcolor{blue!40}{66.33} \\ 
			Fast FMVNet V2    & \cellcolor{blue!40}{\textbf{4.4M}}   & 2.01GB  & 37.7ms  &\cellcolor{blue!40}{26.5}  & \cellcolor{blue!40}{68.55} \\ 
	\end{tabular}}
\end{table}

In Table~\ref{tab:time_kitti_models} and Fig.~\ref{fig:speed_miou_kitti}, we see that the point-based model, WaffleIron (c256) (69.0\%), can achieve a competitive mIoU score with the voxel-based models MinkUNet (cr1.6) (69.41\%) and SPVCNN (cr1.6) (69.06\%). However, for the inference speed, WaffleIron (c256) runs slower than MinkUNet (cr1.6) and SPVCNN (cr1.6) and has only 4.8FPS. Besides, WaffleIron (c256) requires higher GPU memory than other models during inference. We think that too many pointwise operations and the feature transformation between the ``point token-mixing layer" and the ``channel-mixing layer" lead to the low inference speed and high GPU memory. 

The voxel-based models, MinkUNet (cr1.6) and SPVCNN (cr1.6), get the highest and second-highest mIoU scores. Also, the models with sparse convolution operations consume low GPU memory during inference. However, the inference speeds of both models are not high. This is because, except for real computation, other time-consuming operations, such as gathering and scattering data, are required in the sparse convolution operations~\cite{torchsparse23}.

The range image-based models, CENet and Fast FMVNet V2, can run at high speed with competitive performance. Specifically, under the condition of $<10$M model parameters (see the blue masks Table~\ref{tab:time_kitti_models}), Fast FMVNet V2 surpasses MinkUNet (cr0.5) and SPVCNN (cr0.5) in terms of inference speeds and mIoU scores. Also, Fast FMVNet V2 is 5.5 times faster than WaffleIron (c256). CENet obtains the lowest inference latency among the models. However, for the mIoU scores, CENet and Fast FMVNet V2 are inferior to the WaffleIron (c256) and MinkUNet (cr1.6) and SPVCNN (cr1.6). We think that the models not explicitly taking advantage of the depth information and the deformed objects in the range images lead to inferior performance.

\begin{table*}[h]
	\caption{Comparison results on the nuScenes~\cite{nuscenes_panoptic} validation dataset in terms of IoU and mIoU scores (\%). \textbf{Bold font}: the best result; \underline{Underline}: the second-best result.}
	\label{tab:32x1088_nuscenes_val_results}
	\centering
	\scalebox{0.96}{
		\begin{tabular}{l|l|c|c|c|c|c|c|c|c|c|c|c|c|c|c|c|c}
			Models	 &mIoU &\rotatebox{90}{Barrier} &\rotatebox{90}{Bicycle} &\rotatebox{90}{Bus} &\rotatebox{90}{Car} &\rotatebox{90}{Constr. Veh.} &\rotatebox{90}{Motorcycle} &\rotatebox{90}{Pedestrian} &\rotatebox{90}{Traffic Cone} &\rotatebox{90}{Trailer} &\rotatebox{90}{Truck} &\rotatebox{90}{Drive. Sur.} &\rotatebox{90}{Other Flat} &\rotatebox{90}{Sidewalk} &\rotatebox{90}{Terrain} &\rotatebox{90}{Manmade} &\rotatebox{90}{Vegetation}  \\ \hline
			
			WaffleIron (c256) &74.9 &\underline{64.6} &\textbf{51.0} &\underline{94.1} &83.7 &48.8 &\underline{87.1} &63.1 &62.9 &\underline{71.0} &80.7 &96.6 &\underline{73.9} &74.8 &73.4 &87.3 &84.9 \\
			WaffleIron (c384) &\underline{75.6} &\textbf{64.6} &\underline{50.4} &93.0 &85.0 &49.7 &\textbf{87.6} &\underline{67.1} &\textbf{65.0} &\textbf{71.2} &\textbf{83.9} &\underline{96.6} &\textbf{73.9} &\underline{74.8} &73.4 &87.6 &85.2 \\ \hline
			
			MinkUNet (cr2.0)  &70.5 &62.1 &47.4 &93.5 &\underline{87.4} &41.6 &84.9 &40.7 &33.0 &64.7 &82.8 &96.3 &68.6 &74.1 &73.3 &89.1 &87.9 \\
			MinkUNet (cr1.6)  &70.1 &61.2 &47.6 &92.6 &87.3 &39.0 &84.7 &40.0 &30.3 &65.6 &81.5 &96.3 &71.3 &73.9 &73.5 &88.9 &87.7 \\
			MinkUNet (cr1.0)  &68.0 &60.0 &44.3 &92.4 &87.3 &28.8 &83.5 &37.8 &27.5 &61.1 &80.4 &96.0 &67.9 &72.8 &72.7 &88.3 &87.4 \\
			MinkUNet (cr0.64) &66.4 &58.6 &41.6 &90.8 &86.6 &29.9 &80.6 &35.1 &23.0 &54.2 &82.0 &95.6 &67.1 &71.0 &72.4 &87.2 &86.5 \\
			MinkUNet (cr0.5)  &65.1 &56.9 &40.2 &89.0 &86.3 &23.0 &80.4 &33.7 &21.8 &52.8 &80.1 &95.4 &66.7 &70.2 &72.0 &86.6 &86.3 \\ \hline
			
			SPVCNN (cr2.0)    &70.8 &62.3 &48.3 &92.2 &\textbf{87.6} &42.3 &84.5 &42.3 &32.0 &67.8 &83.3 &96.4 &70.8 &74.0 &73.3 &89.1 &87.6 \\
			SPVCNN (cr1.6)    &70.2 &61.4 &46.2 &93.1 &87.0 &39.4 &84.9 &41.6 &28.3 &68.2 &82.7 &96.2 &70.2 &73.8 &73.4 &89.0 &87.7 \\
			SPVCNN (cr1.0)    &68.3 &60.9 &43.9 &91.4 &87.0 &34.8 &83.1 &37.4 &25.2 &61.8 &82.6 &95.9 &68.8 &71.9 &72.2 &88.1 &87.0 \\
			SPVCNN (cr0.64)   &66.1 &58.4 &43.0 &90.5 &86.8 &25.7 &81.0 &36.3 &23.6 &53.2 &80.9 &95.6 &66.6 &71.0 &71.9 &87.2 &86.5 \\
			SPVCNN (cr0.5)    &64.9 &56.5 &41.3 &89.5 &86.4 &25.3 &79.1 &32.5 &21.8 &49.3 &81.0 &95.4 &66.2 &69.6 &71.2 &86.6 &86.0 \\ \hline
			
			CENet             &74.8 &64.2 &44.5 &91.2 &85.7 &\underline{51.5} &85.3 &66.4 &\underline{64.9} &65.9 &79.2 &96.3 &73.7 &74.1 &\textbf{74.9} &\underline{89.9} &\underline{88.3} \\
			Fast FMVNet V2    &\textbf{75.8} &64.4 &42.9 &\textbf{95.2} &87.2 &\textbf{56.2} &86.9 &\textbf{70.8} &62.2 &66.6 &\underline{83.8} &\textbf{96.7} &72.3 &\textbf{75.1} &\underline{74.2} &\textbf{90.1} &\textbf{88.4} \\ 
	\end{tabular}}
\end{table*}

\begin{table}[t]
	\caption{Comparison results among the point cloud segmentation models~\cite{waffleiron23,minkowski2019,spvnas_2020,pdm2024} on the nuScenes~\cite{nuscenes_panoptic} validation dataset in terms of model parameters (Params.), max GPU memory allocated (Mem.), inference latency, frames per second (FPS), and mIoU scores (\%).}
	\label{tab:time_nuscenes_models}
	\centering
	\scalebox{0.95}{
		\begin{tabular}{l|c|c|c|c|c}
			Methods           &Params. & Mem.  & Latency &FPS   & mIoU (\%)  \\ \hline 
			WaffleIron (c256) & \cellcolor{blue!40}{6.8M}  &0.66GB & 51.4ms  & \cellcolor{blue!40}{19.5} & \cellcolor{blue!40}{74.86} \\
			WaffleIron (c384) & 15.1M  &1.00GB & 76.2ms  & 13.1 & 75.57 \\ \hline
			
			MinkUNet (cr2.0)  & 151.5M &0.84GB & 66.4ms  & 15.1 & 70.46 \\ 
			MinkUNet (cr1.6)  & 96.5M  &0.58GB & 59.9ms  & 16.7 & 70.07 \\   
			MinkUNet (cr1.0)  & 37.9M  &0.28GB & 55.2ms  & 18.1 & 68.00 \\  
			MinkUNet (cr0.64) & 15.3M  &0.16GB & 53.7ms  & 18.6 & 66.39 \\ 
			MinkUNet (cr0.5)  & \cellcolor{blue!40}{9.5M}   &\textbf{0.12GB} & 52.0ms  & \cellcolor{blue!40}{19.2} & \cellcolor{blue!40}{65.08} \\ \hline
			
			SPVCNN (cr2.0)    & 151.7M &0.84GB & 68.3ms  & 14.7 & 70.86 \\        
			SPVCNN (cr1.6)    & 96.7M  &0.58GB & 62.5ms  & 16.0 & 70.20 \\ 
			SPVCNN (cr1.0)    & 37.9M  &0.28GB & 55.4ms  & 18.1 & 68.27 \\ 
			SPVCNN (cr0.64)   & 15.3M  &0.16GB & 54.3ms  & 18.4 & 66.13 \\ 
			SPVCNN (cr0.5)    & \cellcolor{blue!40}{9.5M}   &0.12GB & 54.0ms  & \cellcolor{blue!40}{18.5} & \cellcolor{blue!40}{64.86} \\ \hline 
			
			CENet             & \cellcolor{blue!40}{6.8M}   &0.70GB & \textbf{14.2ms}  & \cellcolor{blue!40}{\textbf{70.5}} & \cellcolor{blue!40}{74.76} \\ 
			Fast FMVNet V2    & \cellcolor{blue!40}{\textbf{4.4M}}   &0.55GB & 16.3ms  & \cellcolor{blue!40}{61.5} & \cellcolor{blue!40}{\textbf{75.81}} \\ 
	\end{tabular}}
\end{table}

In Table~\ref{tab:64x2048_kitti_val_results}, we see that the point-based model, WaffleIron, is good at recognizing irregularly shaped or thin objects, such as \textit{Person}, \textit{Other-ground}, \textit{Vegetation}, \textit{Trunk}, \textit{Terrain}, and \textit{Pole}. We think that projecting the objects onto the XY, XZ, and YZ planes in WI blocks enables the model to process these objects easily. By contrast, due to the 3D convolution operations, the voxel-based models, MinkUNet (cr1.6) and SPVCNN (cr1.6) can accurately identify regularly shaped and non-planar objects, such as \textit{Car}, \textit{Bicycle}, \textit{Motorcycle}, \textit{Truck}, \textit{Other-vehicle}, \textit{Bicyclist}, and \textit{Building}. Here, note that \textit{Bicyclist} is seen as the regularly shaped object because it shows fewer gestures than \textit{Person}. Compared with point- and voxel-based models, range image-based models are good at classifying planar objects, such as \textit{Road}, \textit{Parking}, \textit{Sidewalk}, \textit{Fence}, and \textit{Traffic-sign}. The reason is that in the range images, these objects are regularly shaped (see the pink road and purple sidewalk in Fig.~\ref{fig:motion_compensation}(d)). Besides, an interesting thing is that due to the projection in WI blocks, WaffleIron can also recognize these objects well except \textit{Parking}. However, MinkUNet and SPVCNN cannot classify these objects well apart from \textit{Parking} because it is difficult for sparse 3D convolution operations to extract effective features from a planar object.

In summary, range image-based models are suitable for real-time applications. Besides, for safety-critical applications, such as autonomous driving, the combination of voxel- and range image-based models will be a good choice in detecting free spaces and the objects on the road.

\begin{figure}[t]
	\centering
	\includegraphics[width=0.98\columnwidth]{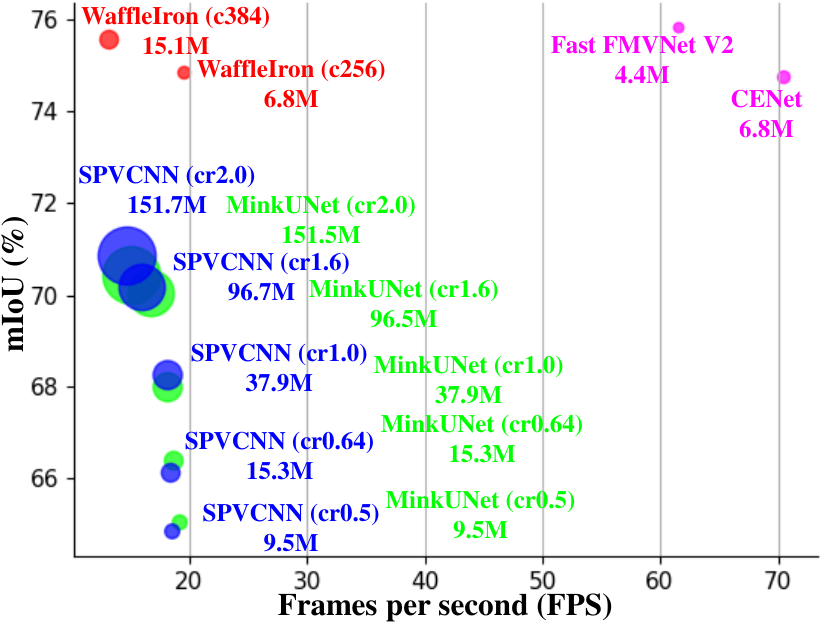}
	\caption{Comparison results among point-, voxel-, and range image-based models~\cite{waffleiron23,minkowski2019,spvnas_2020,pdm2024} on the nuScenes~\cite{nuscenes_panoptic} validation dataset in terms of the model parameters (M), frames per second (FPS), mIoU scores (\%). The sizes of the circles indicate the number of model parameters.}
	\label{fig:speed_miou_nuscenes}
\end{figure}

\subsubsection{Comprehensive Comparisons on nuScenes}
The comprehensive comparison results on the nuScenes dataset are provided in Tables~\ref{tab:time_nuscenes_models} and~\ref{tab:32x1088_nuscenes_val_results} and Fig.~\ref{fig:speed_miou_nuscenes}.

In Table~\ref{tab:time_nuscenes_models} and Fig.~\ref{fig:speed_miou_nuscenes}, we see that compared with voxel-based models MinkUNet and SPVCNN, the point-based model WaffleIron and range image-based models CENet and Fast FMVNet V2 can classify each point well. This is because the nuScenes point clouds are sparser than the SemanticKITTI point clouds and the sparse 3D convolution operations have difficulties to extract effective features from a few points in a local 3D space. By contrast, the range image and the XY, XZ, and YZ planes in WI blocks are dense ``representations" of the point clouds. Therefore, CENet, Fast FMVNet V2, and WaffleIron can easily extract effective features and achieve better performance. Besides, the range image-based model, Fast FMVNet V2, obtains the best mIoU score among all models and runs at high speed (up to 61.5 FPS).

In addition, for MinkUNet and SPVCNN, simply increasing the model capacity by expanding the number of channels, such as MinkUNet (cr2.0) and SPVCNN (cr2.0), can increase the mIoU scores, but the performance gain is limited. Alternatively, enlarging the kernel size in MinkUNet and SPVCNN could increase the performance on the nuScenes dataset, but the large kernel size leads to unacceptable inference latency~\cite{pointtransv32024}.

In summary, for the sparse point clouds, such as nuScenes, the range image-based model should be a good choice regarding the inference latency and mIoU score. 

\section{Conclusion}
In this paper, we provide thorough comparisons from an application view among point-, voxel-, and range image-based models, namely WaffleIron, MinkUNet, SPVCNN, CENet, and Fast FMVNet V2. We first study the impacts of motion compensation and VRCrop on the point- and voxel-based models. Then, comprehensive comparisons among the above models are provided regarding the number of model parameters, max GPU memory allocated during inference, inference latency, frames per second, intersection-over-union (IoU) and mean IoU (mIoU) scores on the SemanticKITTI and nuScenes datasets. In the future, we plan to integrate the point cloud segmentation models to existing simultaneous localization and mapping (SLAM) systems to build semantic maps.

\bibliographystyle{IEEEtran}
\bibliography{main}

\begin{thebibliography}{10}
\providecommand{\url}[1]{#1}
\csname url@samestyle\endcsname
\providecommand{\newblock}{\relax}
\providecommand{\bibinfo}[2]{#2}
\providecommand{\BIBentrySTDinterwordspacing}{\spaceskip=0pt\relax}
\providecommand{\BIBentryALTinterwordstretchfactor}{4}
\providecommand{\BIBentryALTinterwordspacing}{\spaceskip=\fontdimen2\font plus
\BIBentryALTinterwordstretchfactor\fontdimen3\font minus
  \fontdimen4\font\relax}
\providecommand{\BIBforeignlanguage}[2]{{%
\expandafter\ifx\csname l@#1\endcsname\relax
\typeout{** WARNING: IEEEtran.bst: No hyphenation pattern has been}%
\typeout{** loaded for the language `#1'. Using the pattern for}%
\typeout{** the default language instead.}%
\else
\language=\csname l@#1\endcsname
\fi
#2}}
\providecommand{\BIBdecl}{\relax}
\BIBdecl

\bibitem{semantickitti_2019_behley}
J.~Behley, M.~Garbade, A.~Milioto, J.~Quenzel, S.~Behnke, C.~Stachniss, and
  J.~Gall, ``Semantickitti: A dataset for semantic scene understanding of lidar
  sequences,'' in \emph{International Conference on Computer Vision}, 2019, pp.
  9297--9307.

\bibitem{semanticposs_2020}
Y.~Pan, B.~Gao, J.~Mei, S.~Geng, C.~Li, and H.~Zhao, ``Semanticposs: A point
  cloud dataset with large quantity of dynamic instances,'' in \emph{IEEE
  Intelligent Vehicles Symposium}, 2020, pp. 687--693.

\bibitem{nuscenes_panoptic}
W.~K. Fong, R.~Mohan, J.~V. Hurtado, L.~Zhou, H.~Caesar, O.~Beijbom, and
  A.~Valada, ``Panoptic nuscenes: A large-scale benchmark for lidar panoptic
  segmentation and tracking,'' \emph{IEEE Robotics and Automation Letters},
  vol.~7, no.~2, pp. 3795--3802, 2022.

\bibitem{domainAdpt2024}
A.~Xiao, J.~Huang, K.~Liu, D.~Guan, X.~Zhang, and S.~Lu, ``Domain adaptive
  lidar point cloud segmentation via density-aware self-training,'' \emph{IEEE
  Transactions on Intelligent Transportation Systems}, pp. 1--13, 2024.

\bibitem{sa_loam_2021}
L.~Li, X.~Kong, X.~Zhao, W.~Li, F.~Wen, H.~Zhang, and Y.~Liu, ``Sa-loam:
  Semantic-aided lidar slam with loop closure,'' in \emph{International
  Conference on Robotics and Automation}, 2021, pp. 7627--7634.

\bibitem{suma++_2019}
X.~Chen, A.~Milioto, E.~Palazzolo, P.~Giguère, J.~Behley, and C.~Stachniss,
  ``Suma++: Efficient lidar-based semantic slam,'' in \emph{International
  Conference on Intelligent Robots and Systems}, 2019, pp. 4530--4537.

\bibitem{pointnet++_2017}
C.~R. Qi, L.~Yi, H.~Su, and L.~J. Guibas, ``Pointnet++: Deep hierarchical
  feature learning on point sets in a metric space,'' in \emph{Neural
  Information Processing Systems}, vol.~30, 2017.

\bibitem{kpconv_2019}
H.~Thomas, C.~R. Qi, J.-E. Deschaud, B.~Marcotegui, F.~Goulette, and L.~J.
  Guibas, ``Kpconv: Flexible and deformable convolution for point clouds,'' in
  \emph{International Conference on Computer Vision}, 2019, pp. 6411--6420.

\bibitem{minkowski2019}
C.~Choy, J.~Gwak, and S.~Savarese, ``4d spatio-temporal convnets: Minkowski
  convolutional neural networks,'' in \emph{Computer Vision and Pattern
  Recognition}, 2019, pp. 3070--3079.

\bibitem{spvnas_2020}
H.~Tang, Z.~Liu, S.~Zhao, Y.~Lin, J.~Lin, H.~Wang, and S.~Han, ``Searching
  efficient 3d architectures with sparse point-voxel convolution,'' in
  \emph{European Conference on Computer Vision}, 2020, pp. 685--702.

\bibitem{spherical_transformer_2023}
X.~Lai, Y.~Chen, F.~Lu, J.~Liu, and J.~Jia, ``Spherical transformer for
  lidar-based 3d recognition,'' in \emph{Computer Vision and Pattern
  Recognition}, 2023, pp. 17\,545--17\,555.

\bibitem{fidnet_2021}
Y.~Zhao, L.~Bai, and X.~Huang, ``Fidnet: Lidar point cloud semantic
  segmentation with fully interpolation decoding,'' in \emph{International
  Conference on Intelligent Robots and Systems}, 2021, pp. 4453--4458.

\bibitem{cenet_2022}
H.~Cheng, X.~Han, and G.~Xiao, ``Cenet: Toward concise and efficient lidar
  semantic segmentation for autonomous driving,'' in \emph{International
  Conference on Multimedia and Expo}, 2022, pp. 01--06.

\bibitem{filling_missing2024}
B.~Chen, C.~Gong, and J.~Röning, ``Filling missing values matters for range
  image-based point cloud segmentation,'' \emph{IEEE Transactions on
  Intelligent Vehicles}, pp. 1--15, 2024.

\bibitem{pdm2024}
B.~Chen, C.~Gong, A.~Tikanmäki, and J.~Röning, ``Trainable pointwise decoder
  module for point cloud segmentation,'' in \emph{arXiv}, 2024.

\bibitem{loam2014}
J.~Zhang and S.~Singh, ``Loam: Lidar odometry and mapping in real-time,'' in
  \emph{Robotics: Science and systems}, vol.~2, no.~9, 2014, pp. 1--9.

\bibitem{kissicp2023}
I.~Vizzo, T.~Guadagnino, B.~Mersch, L.~Wiesmann, J.~Behley, and C.~Stachniss,
  ``Kiss-icp: In defense of point-to-point icp -- simple, accurate, and robust
  registration if done the right way,'' \emph{IEEE Robotics and Automation
  Letters}, vol.~8, no.~2, pp. 1029--1036, 2023.

\bibitem{openpcseg2023}
Y.~Liu, R.~Chen, X.~Li, L.~Kong, Y.~Yang, Z.~Xia, Y.~Bai, X.~Zhu, Y.~Ma, Y.~Li,
  Y.~Qiao, and Y.~Hou, ``Uniseg: A unified multi-modal lidar segmentation
  network and the openpcseg codebase,'' in \emph{International Conference on
  Computer Vision}, 2023, pp. 21\,662--21\,673.

\bibitem{taseg24}
X.~Wu, Y.~Hou, X.~Huang, B.~Lin, T.~He, X.~Zhu, Y.~Ma, B.~Wu, H.~Liu, D.~Cai,
  and W.~Ouyang, ``Taseg: Temporal aggregation network for lidar semantic
  segmentation,'' in \emph{Computer Vision and Pattern Recognition}, 2024, pp.
  15\,311--15\,320.

\bibitem{rapid25}
L.~Li, H.~P.~H. Shum, and T.~P. Breckon, ``Rapid-seg: Range-aware pointwise
  distance distribution networks for 3d lidar segmentation,'' in
  \emph{European Conference on Computer Vision}, 2025, pp. 222--241.

\bibitem{empir24}
J.~Sun, C.~Qing, X.~Xu, L.~Kong, Y.~Liu, L.~Li, C.~Zhu, J.~Zhang, Z.~Xiao,
  R.~Chen, T.~Wang, W.~Zhang, and K.~Chen, ``An empirical study of training
  state-of-the-art lidar segmentation models,'' in \emph{arXiv}, 2024.

\bibitem{polarmix_2022}
A.~Xiao, J.~Huang, D.~Guan, K.~Cui, S.~Lu, and L.~Shao, ``Polarmix: A general
  data augmentation technique for lidar point clouds,'' in \emph{Neural
  Information Processing Systems}, vol.~35, 2022, pp. 11\,035--11\,048.

\bibitem{lasermix_2023}
L.~Kong, J.~Ren, L.~Pan, and Z.~Liu, ``Lasermix for semi-supervised lidar
  semantic segmentation,'' in \emph{Computer Vision and Pattern Recognition},
  2023, pp. 21\,705--21\,715.

\bibitem{spconv2022}
S.~Contributors, ``Spconv: Spatially sparse convolution library,''
  \url{https://github.com/traveller59/spconv}, 2022.

\bibitem{torchsparse23}
H.~Tang, S.~Yang, Z.~Liu, K.~Hong, Z.~Yu, X.~Li, G.~Dai, Y.~Wang, and S.~Han,
  ``Torchsparse++: Efficient point cloud engine,'' in \emph{Computer Vision and
  Pattern Recognition Workshops}, 2023, pp. 202--209.

\bibitem{pcnn2018}
S.~Wang, S.~Suo, W.-C. Ma, A.~Pokrovsky, and R.~Urtasun, ``Deep parametric
  continuous convolutional neural networks,'' in \emph{Computer Vision and
  Pattern Recognition}, 2018, pp. 2589--2597.

\bibitem{dgcnn2019}
Y.~Wang, Y.~Sun, Z.~Liu, S.~E. Sarma, M.~M. Bronstein, and J.~M. Solomon,
  ``Dynamic graph cnn for learning on point clouds,'' \emph{ACM Transactions on
  Graphics}, vol.~38, no.~5, 2019.

\bibitem{pointtrans2021}
H.~Zhao, L.~Jiang, J.~Jia, P.~Torr, and V.~Koltun, ``Point transformer,'' in
  \emph{International Conference on Computer Vision}, 2021, pp.
  16\,239--16\,248.

\bibitem{kpconvx24}
H.~Thomas, Y.-H.~H. Tsai, T.~D. Barfoot, and J.~Zhang, ``Kpconvx: Modernizing
  kernel point convolution with kernel attention,'' in \emph{Computer Vision
  and Pattern Recognition}, 2024, pp. 5525--5535.

\bibitem{waffleiron23}
G.~Puy, A.~Boulch, and R.~Marlet, ``Using a waffle iron for automotive point
  cloud semantic segmentation,'' in \emph{International Conference on Computer
  Vision}, 2023, pp. 3356--3366.

\bibitem{pointtransv32024}
X.~Wu, L.~Jiang, P.-S. Wang, Z.~Liu, X.~Liu, Y.~Qiao, W.~Ouyang, T.~He, and
  H.~Zhao, ``Point transformer v3: Simpler, faster, stronger,'' in
  \emph{Computer Vision and Pattern Recognition}, 2024.

\bibitem{cylindrical3d2021}
X.~Zhu, H.~Zhou, T.~Wang, F.~Hong, Y.~Ma, W.~Li, H.~Li, and D.~Lin,
  ``Cylindrical and asymmetrical 3d convolution networks for lidar
  segmentation,'' in \emph{Computer Vision and Pattern Recognition}, 2021, pp.
  9934--9943.

\bibitem{squeezeseg}
B.~Wu, A.~Wan, X.~Yue, and K.~Keutzer, ``Squeezeseg: Convolutional neural nets
  with recurrent crf for real-time road-object segmentation from 3d lidar point
  cloud,'' in \emph{International Conference on Robotics and Automation}, 2018,
  pp. 1887--1893.

\bibitem{squeezenet_2016}
F.~N. Iandola, S.~Han, M.~W. Moskewicz, K.~Ashraf, W.~J. Dally, and K.~Keutzer,
  ``Squeezenet: Alexnet-level accuracy with 50x fewer parameters and
  \ensuremath{<}0.5{MB} model size,'' in \emph{arXiv}, 2016.

\bibitem{rangenet++}
A.~Milioto, I.~Vizzo, J.~Behley, and C.~Stachniss, ``Rangenet ++: Fast and
  accurate lidar semantic segmentation,'' in \emph{International Conference on
  Intelligent Robots and Systems}, 2019, pp. 4213--4220.

\bibitem{yolov3_2018}
J.~Redmon and A.~Farhadi, ``Yolov3: An incremental improvement,'' in
  \emph{arXiv}, 2018.

\bibitem{resnet_2016}
K.~He, X.~Zhang, S.~Ren, and J.~Sun, ``Deep residual learning for image
  recognition,'' in \emph{Computer Vision and Pattern Recognition}, 2016, pp.
  770--778.

\bibitem{rangeformer_2023}
L.~Kong, Y.~Liu, R.~Chen, Y.~Ma, X.~Zhu, Y.~Li, Y.~Hou, Y.~Qiao, and Z.~Liu,
  ``Rethinking range view representation for lidar segmentation,'' in
  \emph{International Conference on Computer Vision}, 2023, pp. 228--240.

\bibitem{segformer2021}
E.~Xie, W.~Wang, Z.~Yu, A.~Anandkumar, J.~M. Alvarez, and P.~Luo, ``Segformer:
  Simple and efficient design for semantic segmentation with transformers,'' in
  \emph{Neural Information Processing Systems}, vol.~34, 2021, pp.
  12\,077--12\,090.

\bibitem{rangevit_2023}
A.~Ando, S.~Gidaris, A.~Bursuc, G.~Puy, A.~Boulch, and R.~Marlet, ``Rangevit:
  Towards vision transformers for 3d semantic segmentation in autonomous
  driving,'' in \emph{Computer Vision and Pattern Recognition}, 2023, pp.
  5240--5250.

\bibitem{vit_iclr_2021}
A.~Dosovitskiy, L.~Beyer, A.~Kolesnikov, D.~Weissenborn, X.~Zhai,
  T.~Unterthiner, M.~Dehghani, M.~Minderer, G.~Heigold, S.~Gelly, J.~Uszkoreit,
  and N.~Houlsby, ``An image is worth 16x16 words: Transformers for image
  recognition at scale,'' in \emph{International Conference on Learning
  Representations}, 2021.

\bibitem{convnext2022}
Z.~Liu, H.~Mao, C.-Y. Wu, C.~Feichtenhofer, T.~Darrell, and S.~Xie, ``A convnet
  for the 2020s,'' in \emph{Computer Vision and Pattern Recognition}, 2022, pp.
  11\,966--11\,976.

\bibitem{data_aug2024}
Q.~Zhu, L.~Fan, and N.~Weng, ``Advancements in point cloud data augmentation
  for deep learning: A survey,'' \emph{Pattern Recognition}, vol. 153, p.
  110532, 2024.

\bibitem{fastlio2}
W.~Xu, Y.~Cai, D.~He, J.~Lin, and F.~Zhang, ``Fast-lio2: Fast direct
  lidar-inertial odometry,'' \emph{IEEE Transactions on Robotics}, vol.~38,
  no.~4, pp. 2053--2073, 2022.

\end{thebibliography}

\vfill

\clearpage

\onecolumn

{\appendix
	
	The supplementary materials are summarized as follows:
	
	\begin{itemize}
		\item Details about the data augmentation combination.
		
		\item Visualization of objects on the SemanticKITTI dataset to support the explanations in the section of ``\textit{Comprehensive Comparisons on SemanticKITTI}". 
		
		\item An explanation of why the mIoU scores of the point cloud segmentation models on the nuScenes dataset are a bit low.
		
		\item Experimental results about the nuScenes point clouds sparser than the SemanticKITTI point clouds to support the explanation in the section of ``\textit{Comprehensive Comparisons on nuScenes}". 
	\end{itemize}
	
	\section*{Details about Data Augmentation Combination}
	For VRCrop~\cite{pdm2024} or traditional copy-rotate-paste augmentation~\cite{polarmix_2022}, the chosen classes in SemanticKITTI~\cite{semantickitti_2019_behley} are \textit{Bicycle}, \textit{Motorcycle}, \textit{Truck}, \textit{Other-vehicle}, \textit{Person}, \textit{Bicyclist}, \textit{Motorcyclist}, \textit{Other-ground}, \textit{Trunk}, \textit{Pole}, and \textit{Traffic-sign}. Similarly, the chosen classes in nuScenes~\cite{nuscenes_panoptic} are \textit{Barrier}, \textit{Bicycle}, \textit{Bus}, \textit{Car}, \textit{Construction Vehicle}, \textit{Motorcycle}, \textit{Pedestrian}, \textit{traffic Cone}, \textit{Trailer}, and \textit{Truck}. 
	
	In addition, the probability of VRCrop is set to 0.9. The probabilities of \textit{Random Scaling}, \textit{Random Flipping}, and \textit{Random Rotation} are set to 0.5, 0.9, and 0.9, respectively. The scaling factor for \textit{Random Scaling} is within $\left[0.95, 1.05\right]$.
	
	\section*{Visualization of Objects in SemanticKITTI}
	In Table~\ref{tab:64x2048_kitti_val_results}, we conclude that WaffleIron~\cite{waffleiron23}, MinkUNet~\cite{minkowski2019}, SPVCNN~\cite{spvnas_2020}, CENet~\cite{cenet_2022}, and Fast FMVNet V2~\cite{pdm2024} are good at processing various objects as follows:
	\begin{itemize}
		\item WaffleIron (point-based model) (process irregularly shaped or thing objects):  \textit{Person}, \textit{Other-ground}, \textit{Vegetation}, \textit{Trunk}, \textit{Terrain}, and \textit{Pole}.
		
		\item MinkUNet (cr1.6) and SPVCNN (cr1.6) (voxel-based models) (process regularly shaped and non-planar objects): \textit{Car}, \textit{Bicycle}, \textit{Motorcycle}, \textit{Truck}, \textit{Other-vehicle}, \textit{Bicyclist}, and \textit{Building}.
		
		\item CENet and Fast FMVNet V2 (range image-based models) (process planar objects): \textit{Road}, \textit{Parking}, \textit{Sidewalk}, \textit{Fence}, and \textit{Traffic-sign}.
	\end{itemize}
	Here, we visualize some objects in Fig.~\ref{fig:kitti_objects} for comparisons.
	
	\begin{figure*}[htp]
		\centering
		\includegraphics[width=0.925\columnwidth]{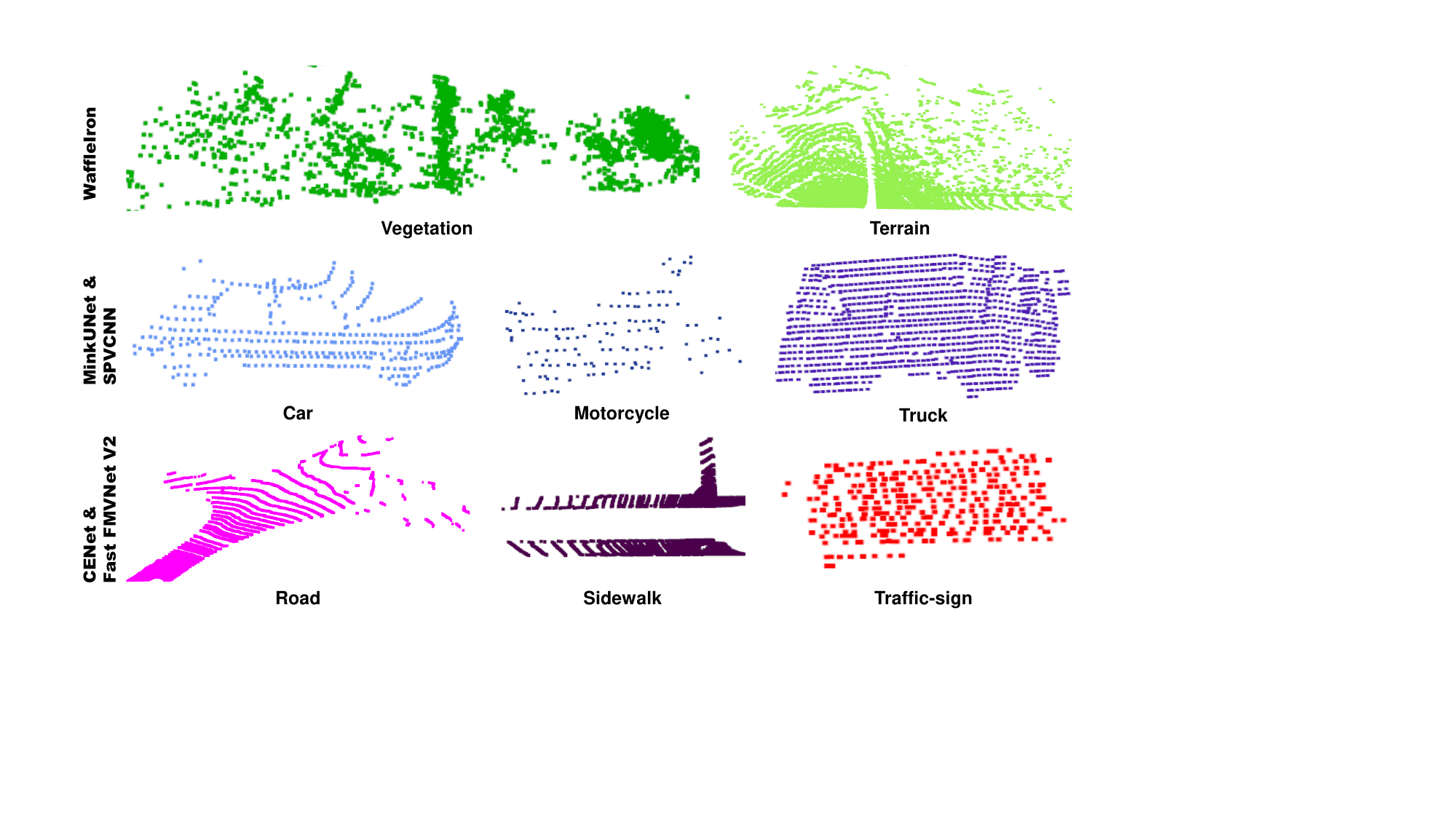}
		\caption{Irregularly shaped objects: \textit{Vegetation} and \textit{Terrain}. Regularly shaped and non-planar objects: \textit{Car}, \textit{Motorcycle}, and \textit{Truck}. Planar objects: \textit{Road}, \textit{Sidewalk}, and \textit{Traffic-sign}.}
		\label{fig:kitti_objects}
	\end{figure*}
	
	In addition, we see that \textit{Bicyclist} is the regularly shaped object while \textit{Person} is not the regularly shaped object. The reason is that \textit{Bicyclist} shows fewer gestures than \textit{Person}, as shown in Fig.~\ref{fig:bicyclist_person}. We see that both hands of a bicyclist should be put on the handlebars, but a person's hands can swing freely.
	
	\begin{figure*}[htp]
		\centering
		\includegraphics[width=0.735\columnwidth]{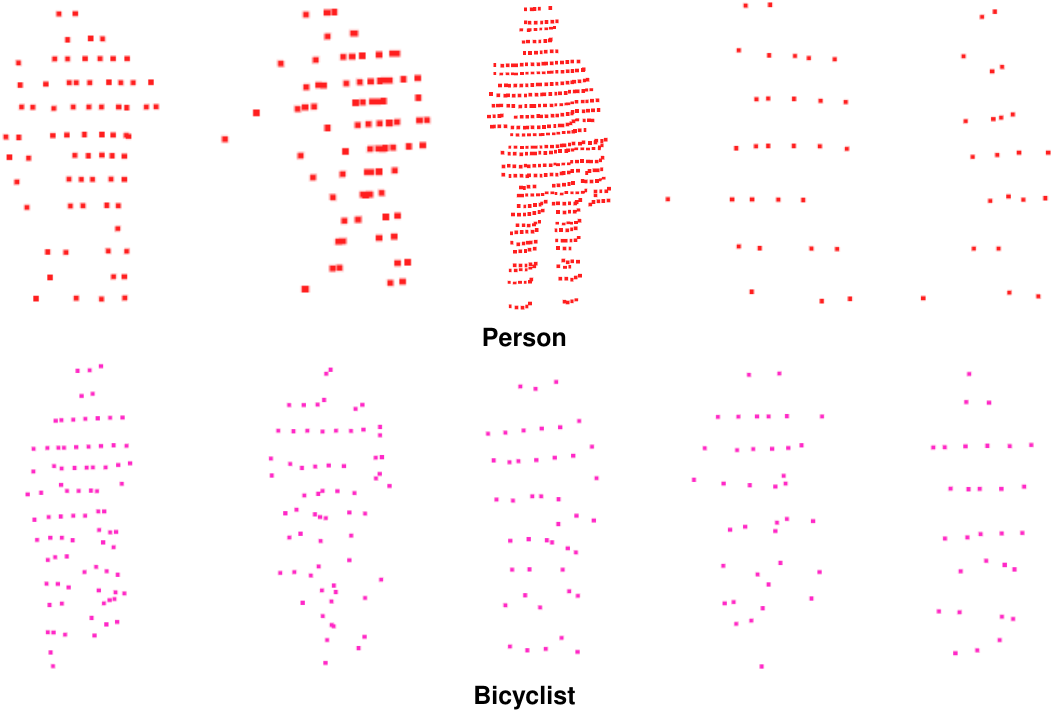}
		\caption{\textit{Person} \textit{vs.} \textit{Bicyclist}. \textit{Bicyclist} shows fewer gestures than \textit{Person} because both hands of a bicyclist should be put on the handlebars of the bicycle, while a person does not care where the hands are.}
		\label{fig:bicyclist_person}
	\end{figure*}

	\section*{Why are the mIoU scores of these models on nuScenes a bit low?}
	We follow the works~\cite{filling_missing2024,pdm2024} to train all models on the nuScenes~\cite{nuscenes_panoptic} dataset. Specifically, erroneously annotated points\cite{lasermix_2023} are removed during training, but the erroneously labelled points are considered during validation. According to the work~\cite{lasermix_2023}, there are at least 5.69\% erroneously annotated points on the nuScenes validation dataset (see Table~\ref{tab:erroneously_points_nuscenes}). Moreover, in practice, we further remove the points within $\left[-0.5m, 0.5m\right]$ during training because these points are normally seen as invalid points (see the issues, https://github.com/ros-drivers/velodyne/issues/61 and https://github.com/nutonomy/nuscenes-devkit/issues/796 and https://github.com/nutonomy/nuscenes-devkit/issues/991). Note that we train all models under the same condition, so the experimental comparisons are still fair. 
	 
	\begin{table}[htp]
		\caption{Percentages of erroneously annotated points on the nuScenes~\cite{nuscenes_panoptic} dataset.}
		\label{tab:erroneously_points_nuscenes}
		\centering
		\scalebox{1.0}{
			\begin{tabular}{c|c|c}
				Ranges                                           &Training Dataset   &Validation Dataset  \\ \hline 
				$\boldsymbol{x} \not\in \left[-50m, 50m\right], \boldsymbol{y} \not\in \left[-50m, 50m\right], \boldsymbol{z} \not\in \left[-5m, 3m\right]$~\cite{lasermix_2023}                  &5.53\%             & 5.69\%     \\ \hline
				+ $\sqrt{\boldsymbol{x}^2 + \boldsymbol{y}^2 + \boldsymbol{z}^2} < 0.5m$                &5.64\%             & 5.86\%     \\
					
		\end{tabular}}
	\end{table}

	\section*{nuScenes point clouds are sparser than SemanticKITTI point clouds.}
	Here, we provide a metric, namely Point Density, to show that the nuScenes point clouds are sparser than the SemanticKITTI point clouds. Specifically, (1) the 3D space within $\boldsymbol{x} \in \left[-50m, 50m\right], \boldsymbol{y} \in \left[-50m, 50m\right], \boldsymbol{z} \in \left[-5m, 3m\right], \text{and} \sqrt{\boldsymbol{x}^2 + \boldsymbol{y}^2 + \boldsymbol{z}^2} > 1m$ is chosen to make sure the space receiving enough points. (2) The 3D space is voxelized with various cubic sizes, such as 0.05$m^3$ and 0.1$m^3$. Here, 0.05$m^3$ is the experimental setting for MinkUNet~\cite{minkowski2019} and SPVCNN~\cite{spvnas_2020} on the SemanticKITTI dataset. 0.1$m^3$ is the experimental setting for both models on the nuScenes dataset. (3) The point density is calculated by the following steps: selecting the top ten voxels with the most points and averaging these voxels to get one point density value for one point cloud, calculating all point density values over all point clouds, and averaging these values to obtain the final point density. Here, choosing the top ten voxels ensures that each voxel contains enough points.
	
	In Table~\ref{tab:nuscenes_sparser_semantickitti}, we see that the point density on the nuScenes dataset is smaller than that on the SemanticKITTI dataset. Besides, with the cubic sizes of $\left\{0.3, 0.4, 0.5, 0.6, 0.7, 0.8, 0.9, 1.0\right\}m^2$, the SemanticKITTI point density is about 4 times of the nuScenes point density. This is because the nuScenes point clouds are collected by a 32-beam LiDAR sensor with 20Hz, and the SemanticKITTI point clouds are captured by a 64-beam LiDAR sensor with 10Hz.  
	
	\begin{table}[htp]
		\caption{The number of points within a voxel (\textit{i.e.}, point density) under different cubic sizes on the SemanticKITTI~\cite{semantickitti_2019_behley} and nuScenes~\cite{nuscenes_panoptic} validation datasets, respectively.}
		\label{tab:nuscenes_sparser_semantickitti}
		\centering
		\scalebox{1.0}{
			\begin{tabular}{l|c|c|c|c|c|c|c|c|c|c|c}
				Datasets                                       &0.05$m^3$ &0.1$m^3$ &0.2$m^3$  &0.3$m^3$ &0.4$m^3$ &0.5$m^3$ &0.6$m^3$  &0.7$m^3$ &0.8$m^3$ &0.9$m^3$ & 1.0$m^3$ \\ \hline 
				SemanticKITTI~\cite{semantickitti_2019_behley} &12.0      &37.9     &119.4     &221.2    &337.8    &475.2    &603.6  &763.8 &896.3 &1050.5 &1190.4        \\ \hline
				nuScenes~\cite{nuscenes_panoptic}              &9.3       &17.8     &39.0      &58.7     &78.0     &105.9    &133.0  &157.8 &194.0 &223.9 &252.5
		\end{tabular}}
	\end{table}
}

\end{document}